\documentclass[onecolumn]{article}
\usepackage{amsmath}
\usepackage{graphicx}
\usepackage{hyperref}       
\usepackage{url}            
\usepackage[english]{babel}
\usepackage[nottoc]{tocbibind}
\usepackage[square,numbers]{natbib}
\usepackage{authblk}
\usepackage{subfigure}
\usepackage{booktabs}
\usepackage{multirow}
\usepackage{threeparttable}
\usepackage{lscape}

\graphicspath{ {./images/} }

\title{\textbf{Dynamic Activation Pitfalls in LLaMA Models: An Empirical Study}}

\author{Chi Ma}
\author{Mincong Huang}
\author{Chao Wang}
\author{Yujie Wang\thanks{Contact email:wangyujie37@meituan.com}}
\author{Lei Yu}
\affil{Meituan}
\date{}
\begin{document}
\maketitle
\begin{abstract}
In this work, we systematically investigate the efficacy of dynamic activation mechanisms within the LLaMA family of language models. Despite the potential of dynamic activation methods to reduce computation and increase speed in models using the ReLU activation function, our empirical findings have uncovered several inherent pitfalls in the current dynamic activation schemes. Through extensive experiments across various dynamic activation strategies, we demonstrate that LLaMA models usually underperform when compared to their ReLU counterparts, particularly in scenarios demanding high sparsity ratio. 
We attribute these deficiencies to a combination of factors: 
1) the inherent complexity of dynamically predicting activation heads and neurons; 
2) the inadequate sparsity resulting from activation functions; 
3) the insufficient preservation of information resulting from KV cache skipping. 
Our analysis not only sheds light on the limitations of dynamic activation in the context of large-scale LLaMA models but also proposes roadmaps for enhancing the design of future sparsity schemes. 
\end{abstract}

\section{Introduction} \label{intro}
Large Language Models (LLMs), due to their immense parameter size, require substantial computations during both training and inference phases. Consequently, reducing computation and inference latency while maintaining model performance has emerged as a critical research direction. To address this challenge, various sparsity techniques have been proposed to decrease the resource requirements of models during runtime by reducing the number of parameters or computations involved.

Sparsity techniques typically encompass two types: static and dynamic. \textbf{Static sparsity} techniques compress the model in post-training phase by pruning a portion of the weights. The downside of this approach is that once the pruning is complete, the pruned parts cannot be recovered, which may lead to a degradation in model performance. 
Recent advances in static sparsity have been marked by works such as Wanda\cite{sun2024simple}, SparseGPT\cite{frantar2023sparsegpt}, and LoRAShear\cite{chen2023lorashear}. These works have pioneered new pruning metrics, more efficient pruning process, or have targeted different components to achieve higher levels of sparsity in LLMs. By doing so, they have managed to maintain a relatively low performance degradation at a reduced cost, setting a new benchmark for model efficiency in the field.
\textbf{Dynamic sparsity} techniques represent a paradigm shift in model efficiency optimization, particularly when contrasted with traditional static counterparts. Where static approaches maintain a fixed set of active parameters or computational units throughout the inference process, dynamic sparsity techniques introduce a level of adaptability that is contingent upon the input data. By dynamically selecting which parameters or units to be activated during model inference, these techniques can tailor the computational load to the unique characteristics of each input, thus achieving heightened computational efficiency.

Mirzadeh et al.\cite{mirzadeh2023relu} elucidates the capacity of the ReLU activation function to introduce sparsity and proposes the concept of dynamic activation. Empirical studies in this work have demonstrated that the choice of activation function does not significantly affect accuracy since GeLU, SiLU, and ReLU all perform with comparable precision. However, ReLU can save approximately 30\% of computational resources by introducing sparsity, hence the paper advocates for a resurgence in the use of ReLU activation functions in LLMs. 
Nevertheless, many modern LLMs, such as LLaMA series and Falcon, have been trained using non-ReLU activations, and retraining from scratch is not cost-effective. Therefore, the consideration of implementing ReLU activation through fine-tuning methods is necessary. Furthermore, after achieving ReLU activation through fine-tuning, the paper suggests reusing activated neurons to generate new tokens, which can reduce FLOPS by about 30\% at the cost of increasing perplexity by 1-4\%. 

\textbf{DejaVu}\cite{liu2023deja} identifies that the sparsity introduced by ReLU can be predicted and, based on this discovery, proposes the first viable dynamic activation predictor scheme. The study empirically demonstrates that dynamic activation can significantly accelerate the inference speed of LLMs that utilize ReLU as the activation function. The literature employs a predictor to forecast the activation of neurons and heads within self-attention and multilayer perceptron (MLP) blocks.
Based on the OPT model, the research finds that the sparsity of attention heads is approximately 80\% (on average, only about 20\% of heads are activated per token), and the sparsity of MLP neurons is around 95\% (on average, only about 5\% of neurons are activated per token). This suggests that utilizing only about 20\% of attention heads and roughly 5\% of MLP neurons can yield results with an increase in perplexity of less than 1\%. Leveraging the OPT model, DejaVu can facilitate a 2-6x acceleration in LLMs inference latency at 75\% sparsity. 

Building upon the DejaVu approach, \textbf{ReLU$^2$}\cite{zhang2024relu2} and \textbf{ProSparse}\cite{song2024prosparse} have induced sparsity in LLMs with non-ReLU activations by finding thresholds and replacing activation functions. ReLU$^2$ presents a systematic framework that examines the sparsity of LLMs from three aspects: the trade-off between sparsity and performance, the predictability of sparsity, and hardware affinity. 
Under this framework, comprehensive experiments were conducted on LLMs using different activation functions (including ReLU, SwiGLU, ReGLU, and ReLU$^2$), demonstrating that ReLU$^2$ outperforms aforementioned traditional activation functions. On a 1.3B small model trained from scratch, ReLU$^2$ achieved an average accuracy of 48.9 at 70\% of sparsity with almost no loss to model performance. 
In a similar vein to ReLU$^2$, ProSparse further improves model sparsity through two operations during fine-tuning: initially replacing the activation function with Shifted-ReLU and subsequently incorporating an L1 regularization term to further increase model sparsity, ultimately achieving only a 1-percent increase in perplexity at approximately 80\% of sparsity.

In a conclusion, recent literature on the exploration of sparsity within LLMs through the dynamic activation perspective have given rise to innovative strategies that aim to enhance computational efficiency without substantially sacrificing model performance. Mirzadeh et al. have revisited the ReLU activation function, uncovering its sparsity-inducing potential. Building on this property, DejaVu has developed a dynamic activation predictor capable of accelerating LLMs' inference by selectively activating subsets of neurons and attention heads. Subsequently, ReLU$^2$ and ProSparse have introduced methods such as threshold truncation and regularization to induce and amplify sparsity within LLMs. They also employ new activation functions and fine-tuning process to enable a broader range of models to exhibit sparsity.

Despite these advancements, the literature reveals critical gaps, such as insufficient experiments with non-ReLU LLaMA models and a lack of detailed analysis on the balance between model performance and sparsity. This paper attempts to address these issues from an empirical research standpoint. It extends the DejaVu approach with a series of experiments on the LLaMA series, uncovering inherent challenges in inducing dynamic activation in non-ReLU models. 
Although this paper provides additional experimental insights into dynamic activation, there remains a need for a deeper understanding of the mechanisms behind sparsity and a lack of clear, quantifiable metrics for comparing sparsity strategies. Future research endeavors will focus on these issues to foster the comprehensive development and effective application of dynamic activation techniques.

\section{Methodology} \label{method}

\subsection{Dynamic Activation in MLPs} \label{methodolgy: mlp}
As previously mentioned in section \ref{intro}, dynamic activation leverages routers or predictors to determine the "importance" values of LLMs heads or weights under different inputs, and then selects different activation strategies based on these values. Given an input y, the formula for LLaMA's MLP block is (Equation \ref{eq: MLP block}):

\begin{equation}
    MLP(y)=W^{out}\left [ \sigma(W^{in}y)\odot (V^{in}y) \right ] 
\label{eq: MLP block}
\end{equation}
, where the output of the i-th neuron can be defined as Equation \ref{eq: ith neuron} :

\begin{equation}
    n_i(y)=\left [ \sigma (W^{in}_{i,:}y)\odot (V^{in}_{i,:}y) \right ]W^{out}_{:,i} 
\label{eq: ith neuron}
\end{equation}

\begin{figure}[h]
\centering
\includegraphics[height=0.5\textwidth]{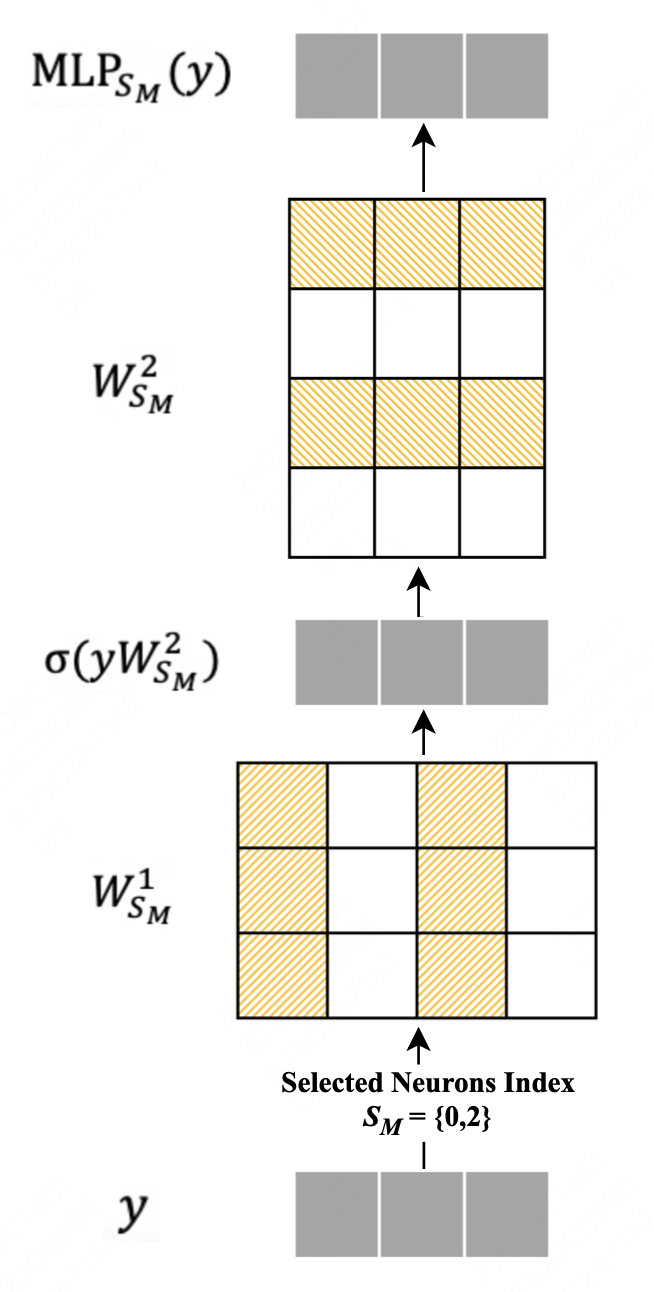}
\caption{Sparsity of MLP block with Neuron \#0 and \#2 be activated}
\label{figure: mlp dynamic activation}
\end{figure}

Building upon the symbolic framework presented in DejaVu, Figure \ref{figure: mlp dynamic activation} provides a visual representation of the aforementioned computation process. Let y denote the input to the MLP block, exemplified here by neurons with index of 0 and 2 are activated. The shaded areas of two weight matrices $W_{S_M}$ represent the computations corresponding to the activated neurons, where $W^{1}_{S_M}$ and $W^{2}_{S_M}$ correspond to $W^{in}$ and $W^{out}$ in Equation \ref{eq: MLP block} and \ref{eq: ith neuron}, respectively.

Following the idea from ReLU$^2$, a threshold truncation method is employed to get activation index set: a threshold \(\epsilon \) is specified, and neurons with magnitudes less than the threshold are treated as zero to induce sparsity. To measure the effects of different thresholds, the literature introduces the concept of CETT (cumulative errors of tail truncation) to assess the impact of sparsity induced by threshold truncation on the MLP output. According to the computation logic of the MLP block, we have (Equation \ref{eq: MLP sum from neuron}):

\begin{equation}
    MLP(y)= {\textstyle \sum_{i=1}^{d_{h}}n_i(y)} 
\label{eq: MLP sum from neuron}
\end{equation}
, where $d_{h}$ is the dimension of the hidden layer in MLP block, hence the formula for CETT is as follows in Equation \ref{eq: mlp cett}:

\begin{equation}
    CETT(y)=\frac{|| {\textstyle \sum_{i\in \mathcal D}n_i(y)} ||_2}{||MLP(y)||_2} ,\mathcal D=\left \{i|\;||n_i(y)||_2<\epsilon   \right \} 
\label{eq: mlp cett}
\end{equation}
, where \(\epsilon \) represents the threshold, $\mathcal D$ is the set of neurons with magnitudes less than the threshold \(\epsilon \), and $n_i$ denotes the output of the i-th neuron from Equation \ref{eq: ith neuron}.

The meaning of CETT is the ratio of the L2 norm of the sum of outputs from neurons that are not activated by token $y$ to the L2 norm of the MLP output. Following ReLU$^2$, we firstly set an upper limit for the CETT and subsequently identify the maximum output magnitude threshold \(\epsilon \) that results in a CETT below the predetermined upper bound. The magnitudes involved in calculating the truncation threshold, here we use L2 norm, can also be computed using different methods. The existing body of research, in conjunction with our experimental results, suggests that a CETT value of 0.2 is the best choice.

The empirically optimal threshold has been established at $CETT=0.2$, as experimental evidence indicates that model performance remains stable when the threshold-determined model sparsity ratio is increased up to this point. In a short word, model performance is relatively stable until it reaches the point of $CETT=0.2$, beyond which there is a significant decline. Furthermore, CETT functions as a performance metric, allowing for increased sparsity ratio without compromising model performance, thus improving computational efficiency. The model's performance is not significantly affected by the truncation of tail neurons when $CETT \le 0.2$. Consequently, $CETT=0.2$ constitutes a balance point that optimizes the model's sparsity activation and computational efficiency without a significant loss in performance.

\subsection{Head Mask and KV Cache Skipping in Attentions} \label{methodology: att}
Self-attention is a crucial element in LLaMA models, yet it represents a considerable source of compute cost (approximately 30\%\cite{ye2024chunkattention}) and latency during the inference process. As mentioned in section \ref{intro}, for a given input token, only a few heads perform significant calculations. 
Existing research on sparsity in attention blocks mainly aims to cut down the number of heads involved in computations, yet the approach to handle the Key-Value (KV) cache is still in early stages without thorough experimental backing or ablation studies.

Similar to the MLP block sparsity in section \ref{methodolgy: mlp}, quickly selecting attention heads without full computation is key to reducing end-to-end latency. For multi-head attention blocks, as shown in Figure \ref{figure: att dynamic activation}, the output of each head is a matrix, thus we only need to calculate the L2 norm of this output matrix and then select heads with larger L2 norms out of all heads. The principle of attention head dynamic activation is illustrated in Figure \ref{figure: att dynamic activation}:

\begin{figure}[h]
\centering
\includegraphics[height=0.5\textwidth]{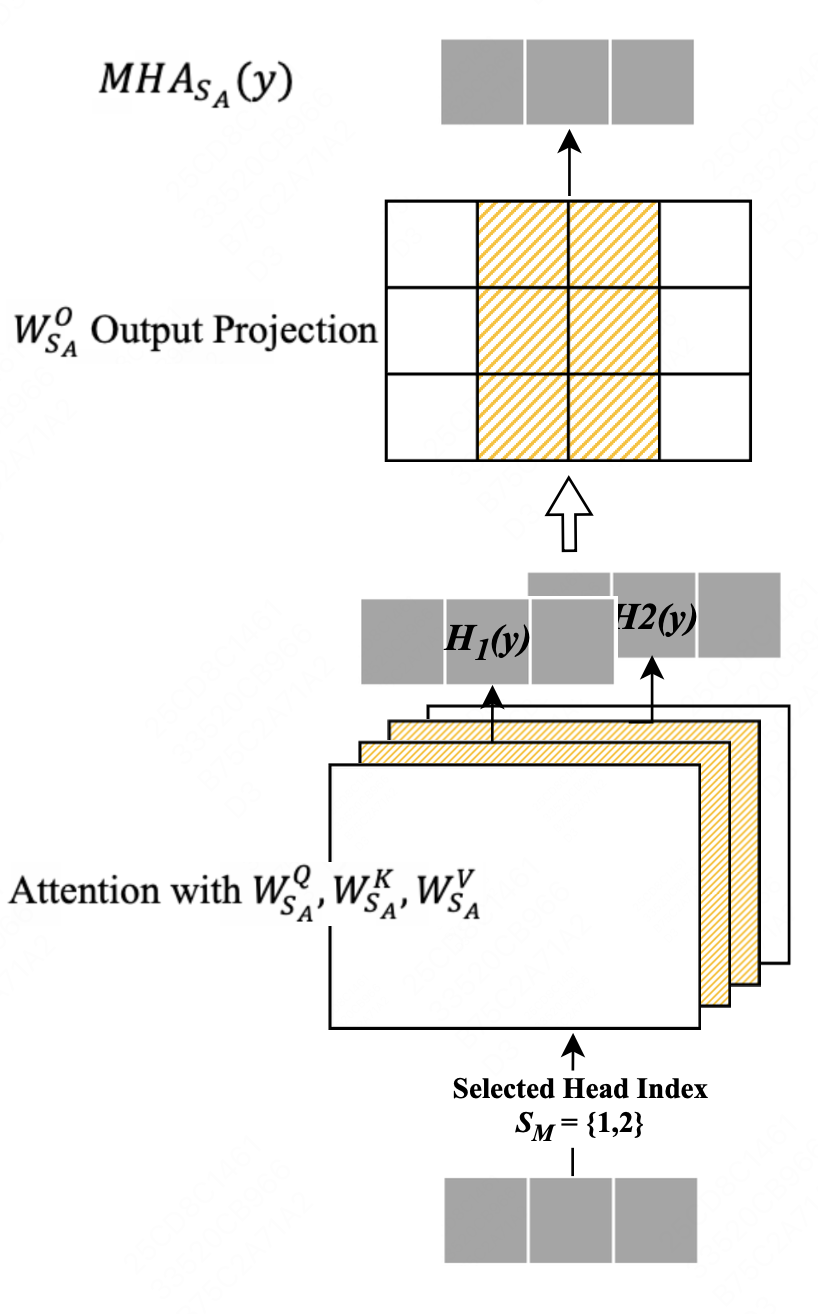}
\caption{Sparsity of Attention block with Head \#1 and \#2 be activated}
\label{figure: att dynamic activation}
\end{figure}

Let $i$ denotes head index, $h$ denotes total head number. then we can get Equation \ref{eq: att three projection}: 

\begin{equation}
    Q_i = QW_i^Q, K_i = KW_i^K, V_i = VW_i^V
    \label{eq: att three projection}
\end{equation}
,where \(0 \le i \le h\).
Based on Figure \ref{figure: att dynamic activation}, suppose we have (Equation \ref{eq: head mask}):

\begin{equation}  
mask_i=\left\{
		\begin{array}{lcl}
			1,  &     & if \; i \in S_A,\\
	        0,  &     & otherwise\\
		\end{array} \right.
\label{eq: head mask}
\end{equation} 
,where (Equation \ref{eq: att active head defination}):
 

\begin{equation}
    S_A=\left \{ i|\;||\;softmax\left(\frac{Q_iK_i^T}{\sqrt{d_k}}\right)V_i\;||_2 > \epsilon \right \} 
    \label{eq: att active head defination}
\end{equation}

Equation \ref{eq: att active head defination} defines an active head. In Figure \ref{figure: att dynamic activation}, the active heads' indices are 1 and 2. The number of heads selected is constrained by CETT. 
In determining the sparsity of the attention block, we adhere to the selection rationale for CETT outlined in the section \ref{methodolgy: mlp} and similarly adopt 0.2 as the CETT value. In this section, we compute the threshold \(\epsilon\) and utilize it to dynamically identify the heads that should be pruned for the current token. Alternatively, we also employ the topK strategy as the second strategy to select the heads by specifying the number of attention heads activated per layer.

In accordance with the calculation process of MLP CETT in Equation \ref{eq: mlp cett}, we obtain CETT for attention block as in Equation \ref{eq: att cett}:

\begin{equation}
    CETT(y)=1 - \frac{|| \text{Concat}_{0 \le i \le h}(mask_i \circ \text{head}_i)W^O ||_2}{||MHA(y)||_2}
    \label{eq: att cett}
\end{equation}
, where \(\epsilon \) represents the threshold, and $h$ is the total number of heads, with $I + S_A = h$.


Upon integrating dynamic sparsity into attention heads, this section turns to the Key-Value (KV) cache in attention blocks, acknowledging its vital role and exploring the possibilities for optimization in models with sparsity. The KV cache is an essential element of the generative mechanism within attention blocks. It preserves the intermediate keys and values, thereby minimizing redundant computations, conserving computational resources, and hastening the inference process. 

As the size and complexity of LLMs continue to grow, the KV cache's role becomes even more pronounced. In the context of models with tens or hundreds of billions of parameters, the KV cache can also become a bottleneck in terms of computation. Thus, optimizing the KV cache not only involves skipping its computation but also rethinking the skipping strategy to ensure that the most relevant information is retained and less critical data is dropped.

Direct pruning of the KV cache might seem like an uncomplicated approach, but finding a way to diminish the computational demands of the KV cache in LLMs without degrading accuracy is a sophisticated and formidable challenge. This balance is particularly delicate because the KV cache is directly tied to the model's ability to quickly access historical information, which is crucial for maintaining context and coherence in generated text.

\begin{figure}[h]
\centering
\includegraphics[height=0.45\textwidth]{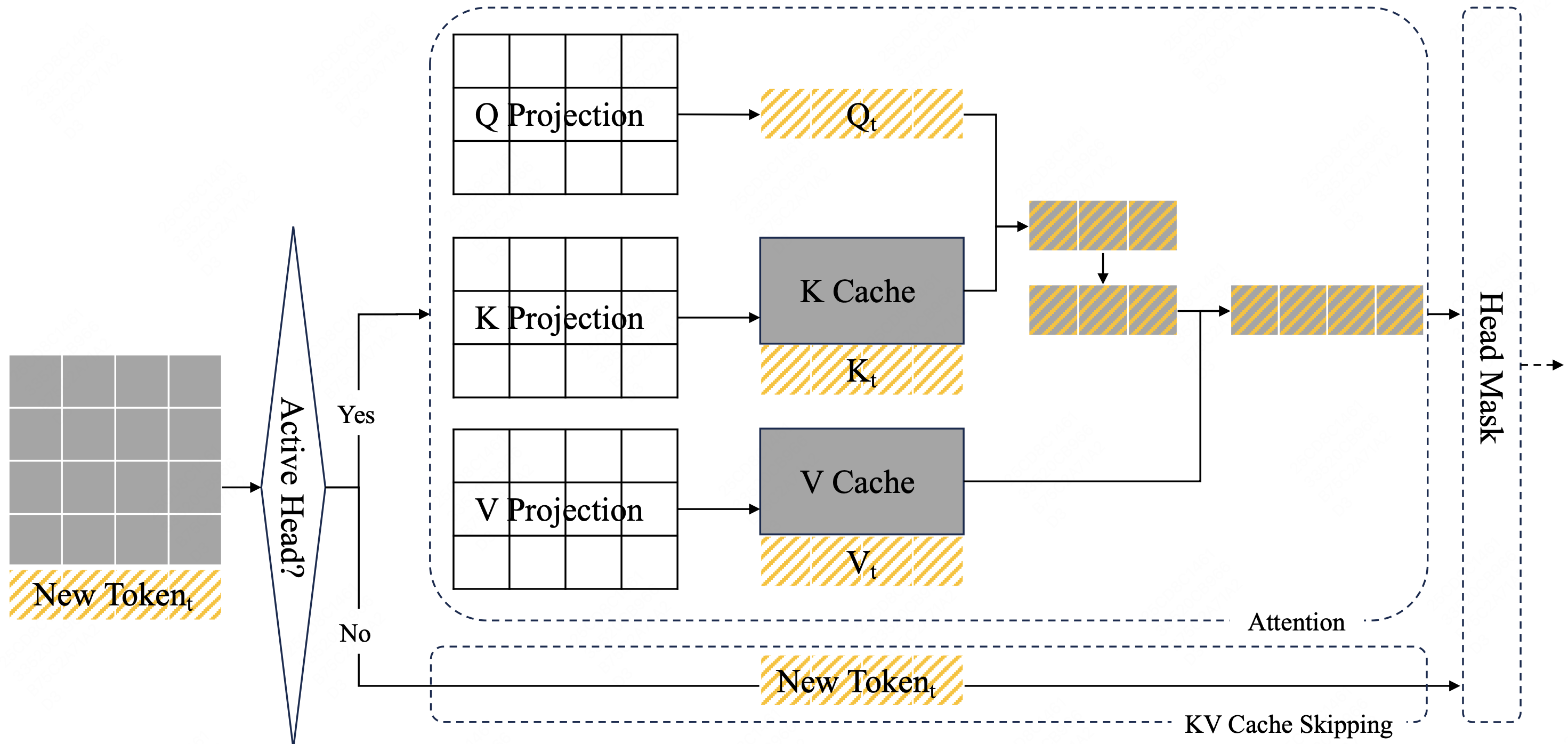}
\caption{KV Cache Skipping}
\label{figure: kv skipping}
\end{figure}

The KV cache strategy we used in this paper comes from DejaVu and aims to reduce computational costs and enhance inference speed by omitting the K and V projections during the KV cache generation. For a visual explanation, please see the Figure \ref{figure: kv skipping}. Specifically, we first identify the active and inactive attention heads within current attention block by using a threshold cut-off or a head predictor. For each active head, we compute the KV cache as usual. However, for heads that are inactive in the current attetion block, we bypass the KV cache computation and instead directly use the input of the current attention block as the KV cache value. This method is designed to lighten the computational burden inherent in the multi-head attention mechanism.

Take threshold-determined head mask in attention blocks as an example. First, we calculate the threshold \(\epsilon \) when $CETT=0.2$, then use the threshold to dynamically determine the heads to be reduced for the current token. For the heads that need to be retained, compute the corresponding KV and store it in the KV cache; also, save a copy of the current token embedding for all other un-selected heads, so that during the future token generation process, if the selected heads lack KV cache, the stored token embedding can be loaded and used to compute KV together.

Expanding on this, the rationale behind skipping the KV projection for inactive heads is rooted in the observation that not all heads contribute equally to the model's performance. By selectively calculating the KV pairs only for those essential heads, we can allocate computational resources more efficiently. This targeted approach not only streamlines the attention mechanism but also opens up the possibility for dynamic resource allocation based on the demands of specific data inputs or even task difficulties. 

Furthermore, this selective KV cache strategy could potentially lead to a more adaptive model architecture. By monitoring the performance impact of different heads over time, the model could learn to activate or deactivate heads in a context-dependent manner, thus optimizing its structure for various linguistic tasks. This dynamic adjustment could result in models that are not only faster but also more accurate, as they would be refined to the intricacies of the task at hand.

\subsection{Sparsity Predictors}
To harness the previously mentioned sparsity for boosting inference speed, we need a method that can quickly and accurately predict which heads and MLP neurons are "active" for a given input. The DejaVu approach utilizes two linear models to make these predictions. The first model determines the attention heads to be "activated" within the attention blocks, while the second model identifies the "efficient" neurons within the MLP blocks.

\begin{figure}[h]
\centering
\subfigure[Serial DejaVu]{
\label{serial dejavu}
\includegraphics[height=0.5\textwidth]{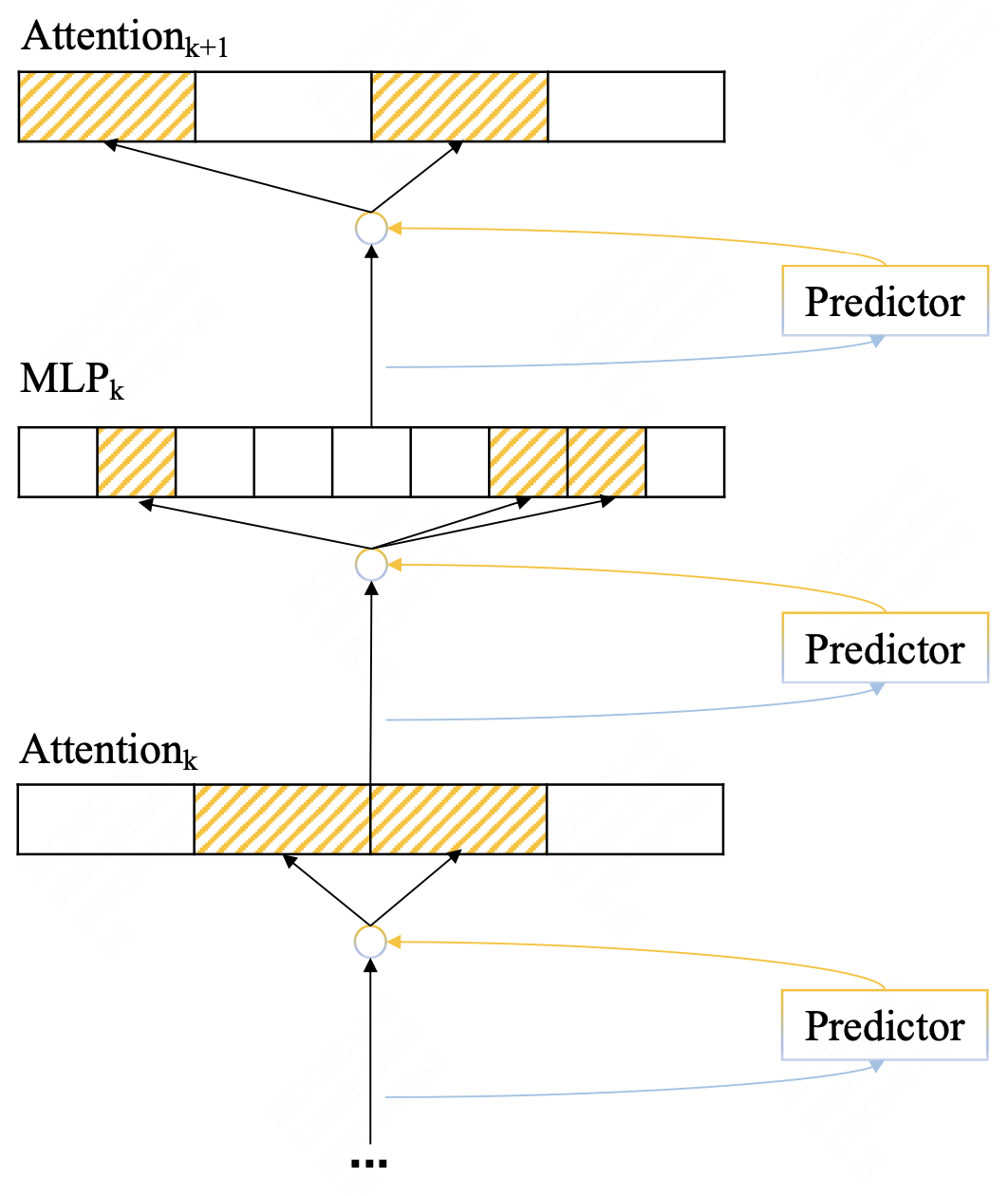}}\subfigure[Parallel DejaVu]{
\label{parallel dejavu}
\includegraphics[height=0.45\textwidth]{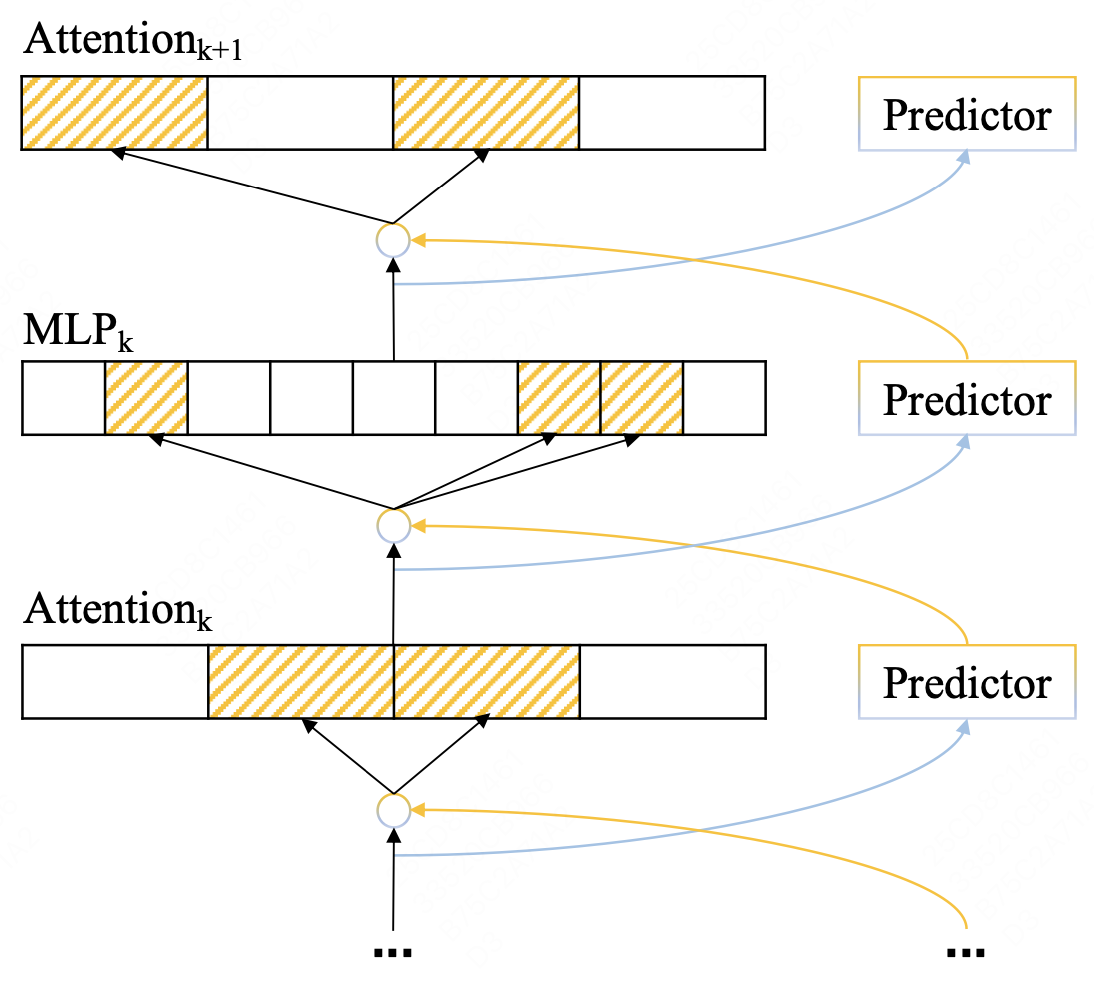}}
\caption{Serial and parallel implementation of DejaVu}
\label{figure: dejavu}
\end{figure}

To illustrate with the prediction of attention heads, let's assume there are 32 heads. The output layer of the predictor thus has 32 dimensions, and it employs a sigmoid function for binary classification, which labels each head as either "active" or "inactive". Training data for this model is sourced from the original, densely connected large language model. During the inference phase, the LLMs' attention inputs and outputs are logged. We compute the L2 norm for each head and then categorize them as positive or negative examples based on a predetermined L2 norm threshold. The concept for selecting neuron indices in the MLP block is the same to that of the attention heads. The implementation of the DejaVu approach, taking a LLaMA module as an example, is depicted in Figure \ref{figure: dejavu}.

As can be seen from Figure \ref{figure: dejavu}, the dual-predictor approach proposed in the DejaVu achieves dynamic activation prediction of attention heads and MLP neurons by adding predictors to the original model architecture.Assume the current module is the L-th layer of the network, and its two sparsity prediction models are denoted as $SP_{Att}^{L}$ and $SP_{MLP}^{L}$, where $SP$ stands for Sparsity Predictor, with the subscript $Att$ indicating prediction for the attention block, and the subscript $MLP$ indicates prediction for the MLP block. The superscript $L$ indicates the current position at the L-th layer of the large model. 

Assuming the current input is $y$, first we use $SP_{Att}^{L}$ to predict the indices of the heads to be selected: \(a=SP_{Att}^L(y)\). Here, $a$ represents the set of indices of the selected heads. Then we calculate the output of attention block that only passes through the heads in $a$: \(out_{Att}=MHA_a(x)\). Next, use the output of attention blocks to predict the indices of the neurons required in MLP blocks, and then calculate the output of the MLP block. 

To enhance the predictive efficiency, we have also incorporated input pre-positioning experiments on LLaMA series of the DejaVu parallel design. These experiments involve testing alternative sources for the MLP predictor's input, shifting from the current layer's MLP input to either: 
1) the previous layer's MLP input; or 
2) the current layer's attention input.

Our experiments reveal that substituting the MLP predictor's input with the current layer's attention block input does not significantly affect the predictor's recall and sparsity. However, using the input from the previous layer's MLP markedly impacts the predictor's performance. To keep hardware efficiency, we recommend that atention predictor utilizes the input from the previous layer's MLP, whereas the MLP Predictor employs the input from the attention block of the same layer.

\section{Empirical Evaluation and Analysis}

In this section, we investigate the performance of dynamic activation in non-ReLU models. Our experiments begin by applying the DejaVu and ReLU$^2$ schemes to thin out the LLaMA-2-7B and LLaMA-3-8B models using a mixed token set derived from Wikipedia. Subsequently, we evaluate the sparsely activated models using the \textbf{lm-evaluation}\cite{eval-harness} framework, comparing their performance against that of the fully dense LLaMA series models. The evaluation metrics encompass model type, predictor type, and performance on linguistic tasks.

The proposed DejaVu framework is expected to achieve significant reductions (around 70\%-95\%) in activated parameters while maintaining or improving upon the linguistic performance of LLMs. This will make LLMs more accessible for deployment on computation-constrained platforms and for use in real-time applications.

\subsection{Threshold Truncation in MLPs} \label{subsec: threshold truncation in mlps}
Following the literature's scheme, experiments were conducted on the LLaMA-2-7B and LLaMA-3-8B models using $2^{17}$ tokens from the Wikipedia dataset. By setting a series of CETT values, the achieved average sparsity and corresponding threshold epsilon are shown in Table \ref{table: cett sparsity and uni-thresholds}. It is observed that at \(CETT=0.2\), the average sparsity under a model-wise universal threshold is only 57.13\% on LLaMA-2-7B and 67.29\% on LLaMA-3-8B. Significant drops in MMLU and CEVAL points are witnessed as the increase of CETT values while using universal thresholds across all layers. We also observed that under a universal threshold, the sparsity gradually decreases from the first layer to the back, with the first layer exceeding 99\% sparsity while the last layer is only slightly over 10\%. 

\begin{table}[h]
    \centering
    \scalebox{0.8}{
    \begin{tabular}{l|l|ccccccc}
        \toprule
        \multicolumn{2}{c|}{\textbf{CETT}} & \textbf{0.01} & \textbf{0.02} & \textbf{0.04} & \textbf{0.1} & \textbf{0.2} & \textbf{0.4} & \textbf{0.5} \\
        \midrule
        \multirow{2}*{\textbf{LLaMA-2-7B}} & Sparsity(\%) & 11.49 & 16.58 & 23.95 & 39.55 & 57.13 & 79.25 & 86.62 \\
        ~ & Threshold & 0.0018 & 0.0032 & 0.0061 & 0.0158 & 0.0349 & 0.0871 & 0.1260 \\
        \midrule
        \multirow{2}*{\textbf{LLaMA-3-8B}} & Sparsity(\%) & 14.10 & 21.04 & 30.72 & 49.08 & 67.29 & 86.91 & 92.47 \\
        ~ & Threshold & 0.0018 & 0.0032 & 0.0061 & 0.0158 & 0.0349 & 0.0871 & 0.1260 \\
        \bottomrule
    \end{tabular}
    }
    \caption{Sparsity and universal thresholds under different CETTs}
    \label{table: cett sparsity and uni-thresholds}
\end{table}

The aforementioned scenario indicates that there is a significant variance in sparsity across different layers, and a universal threshold should not be applied. Hence, using a mixed dataset of $2^{17}$ tokens, layer-specific thresholds were searched for each layer of the LLaMA-2-7B and LLaMA-3-8B models, and the thresholds computed for different CETT values are presented at Table \ref{table: cett layer wise} in Appendix \ref{app: app table}. 

\begin{table}[h]
    \centering
    \scalebox{0.7}{
    \begin{tabular}{l|c|c|ccccc}
    \toprule
    \textbf{Model} & \textbf{CETT} & \textbf{Sparsity} & \textbf{MMLU} & \textbf{TruthfulQA} & \textbf{Winogrande} & \textbf{GSM8K} & \textbf{Average} \\
    \midrule
    \multirow{2}*{\textbf{LLaMA-2-7B}} & 0 & 0 &45.83 & 61.04 & 74.11 & 13.95 & 48.73 \\
    ~ & 0.2 & 67.12\% & 45.62 & 60.66 & 73.88 & 13.65 & 48.45 \\
    \midrule
    \multirow{2}*{\textbf{LLaMA-3-8B}} & 0 & 0 & 66.60 & 56.11 & 76.64 & 49.13 & 62.12 \\
    ~ & 0.2 & 45.84\% & 63.89 & 55.64 & 75.37 & 44.66 & 59.89 \\
    \bottomrule
    \end{tabular}
    }
    \caption{Comparative evaluation of model performace across different CETTs}
    \label{table: cett lm-eval}
\end{table}

By truncating the computed results of the MLP block's input using the layer-specific thresholds obtained at a CETT of 0.2, the lm-eval results are as shown in the Table \ref{table: cett lm-eval}. It can be observed that after truncation using layer-specific thresholds, the model performance is almost lossless with a sparsity of approximately 67\% for LLaMA-2-7B and 45\% for LLaMA-3-8B.

\subsection{Head Mask in Attentions}

In this section, we try to delineate the experimental findings on MHA mechanism of LLaMA-2-7B models, which illuminate the viability and effectiveness of threshold-driven sparsity within attention architectures, contributing to the broader discourse on model pruning within the field of machine learning.

In section \ref{subsec: threshold truncation in mlps}, we reported a series of empirical experiments into the sparsification of neurons within the MLP block, employing a threshold truncation approach. The empirical evidence indicated that such a method is capable of creating over 67\% and 45\% of sparsity in the MLP block while maintaining model performance with negligible degradation. This subsection delves into extending selecting strategy defined by threshold to the attention block, via following the roadmap of MLP block: 
1) the sparsification of attention heads via mask defined by threshold, 
and 2) an examination of KV cache skipping based on head mask.

We first report the result of head sparsification via mask defined by threshold in Table \ref{table: head thresholding}. Then, multiple experiments on KV cache skipping are conducted, and results are reported in Table \ref{table: kv cache gms8k}.

\subsubsection{Threshold-determined head mask in Attentions} \label{head truncation}
In Table \ref{table: head thresholding}, we present a comparative evaluation of the LLaMA-2-7B model's attention block performance post-sparsification of heads under the condition of a CETT set at 0.2. The findings reveal that head sparsification within the attention block, solely utilizing a threshold-defined head mask strategy, can yield approximately 44\% sparsity with minimal impact on the overall efficacy of the model. 

\begin{table}[h]
    \centering
    \scalebox{0.75}{
    \begin{tabular}{l|c|ccccc}
    \toprule
    \textbf{Block name(s)} & \textbf{Sparsity} & \textbf{MMLU} & \textbf{TruthfulQA} & \textbf{Winogrande} & \textbf{GSM8K} & \textbf{Average} \\
    \midrule
    \textbf{Attention} & 44\% & 44.63 & 61.65 & 73.24 & 10.38 & 47.48 \\
    \textbf{MLP \& Attention} & 67\% \& 44\% & 44.55 & 61.29 & 72.45 & 9.62 & 46.98 \\
    \bottomrule
    \end{tabular}
    }
    \caption{Comparative evaluation of LLaMA blocks at $CETT=0.2$}
    \label{table: head thresholding}
\end{table}

When combined with the MLP threshold truncation dynamic activation method previously mentioned in section \ref{subsec: threshold truncation in mlps}, the model exhibits only a marginal decrease in performance across various tasks. These experiments collectively suggest that the threshold-determined truncation or mask approach, when applied to both MLP and attention blocks, can achieve a significant level of sparsity without substantially compromising the model's performance. These insights offer valuable guidance for further model selection and optimization efforts within the realm of machine learning model pruning.

\subsubsection{KV Cache Skipping Based on Head Mask in Attentions}
Key-Value (KV) Cache helps the model by storing the matching pairs of keys and values in the attention mechanism, which cuts down on unnecessary repeat calculations and speeds things up. 
In section \ref{methodology: att}, we delineated the methodology for executing Key-Value (KV) cache skipping, contingent upon the head activation outcomes. However, the empirical evidence of KV cache skipping in current literature remains a blur. This section is dedicated to presenting the experimental findings associated with KV cache skipping after threshold-determined head mask in attention blocks. Additionally, we also elucidate the performance outcomes derived from various skipping strategies.

The DejaVu approach to handling the KV cache poses considerable difficulties in evaluating model performance. Since the mainstream tasks in existing lm-eval are choice questions that generate at most one token, pruning the KV cache is ineffective in this scenario. Therefore, we:
1) switch the evaluation task from choice questions to generative tasks; 
2) conduct ablation experiments on different skipping strategies to determine the effects and roles of KV cache.

\begin{table}[h]
\begin{threeparttable}
    \centering
    \scalebox{0.9}{
    \begin{tabular}{l|cc}
    \toprule
    \textbf{Sparsity type} & \textbf{Strict match} & \textbf{Flexible extract}\\
    \midrule
    \textbf{Dense} & 0.1357 & 0.1395 \\
    \textbf{Head Mask Only} & 0.1137 & 0.1198 \\
    \midrule
    \textbf{KV Skipping} & 0.0008 & 0.0031 \\
    \textbf{Only Skip K} & 0.0045 & 00197 \\
    \textbf{Only Skip V} & 0.0136 & 0.0281 \\
    \textbf{KV Skipping$_w/$ Layer filter} & 0.0728 & 0.0788 \\
    \textbf{Only Skip V$_w/$ Layer filter} & \underline{0.1001} & \underline{0.1039} \\
    \bottomrule
    \end{tabular}
    }
    \begin{tablenotes}\footnotesize
    \item \textbf{Notes}: The term "Dense" refers to the absence of threshold-determined head mask and KV cache skipping in attention blocks. All KV cache skipping approaches are applied after threshold-determined head mask. "KV Skipping", known as the DejaVu KV cache skipping method, is applied to each masked head within an attention block across all layers. In contrast, " Onlly Skip K" focuses solely on managing the key cache following the mask of heads, , Keep V cache untouched. Meanwhile, "Only Skip V" specifically addresses the processing of the V cache for each truncated head. The "Layer filter" refers to a selective application of cache skipping strategies, targeting only those layers with a sparsity greater than 50\%. The \underline{underlined} numbers are the optimal result over all kv cache skipping strategies.
    \end{tablenotes}
    \caption{LLaMA-2-7B performance on GSM8k under various sparsity strategies}
    \label{table: kv cache gms8k}
\end{threeparttable}
\end{table}

The experimental results in Table \ref{table: kv cache gms8k} indicate that merely performing threshold-determined head mask in attention blocks does not significantly affect the model evaluation results. However, the treatment of the KV cache proposed in DejaVu results in a substantial loss of evaluation effectiveness unless cache skipping is applied only to layers with sparsity greater than 50\% for the V cache, without processing K. 

The reason lies in the self-attention mechanism, where the Query (Q), Key (K) and Value (V) all originate from the same input sequence. After K undergoes a linear transformation, it is matched with Q to calculate the degree of alignment, and based on this, attention weights are allocated to different V. This process can be viewed as weighting information so that information more relevant to the current task receives more focus. Through this mechanism, K helps in filtering and selecting pertinent information from the input sequence. Q determines the importance of information through interaction with K, thereby achieving a weighted summation of V. However, skipping the K cache is equivalent to bypassing the linear transformation of K and directly calculating the match with Q, resulting in a significant deviation from the results of the dense activated attention mechanism.

\subsection{Predictors and Trade-off}
In Section \ref{method}, we discussed the integration of predictors into the MLP and attention blocks to facilitate dynamic sparsity. This subsection extends that approach by incorporating predictors into both blocks. Our experimental analysis reveals that for both the attention and MLP blocks of the vanilla LLaMA model, employing a simple two-layer linear predictor results in a significant trade-off between the predicted sparsity and overall model performance. We discovered that increasing the complexity of the predictor's structure can overcome this trade-off. 

\subsubsection{Predictors in MLPs}
Table \ref{table: mlp predictor with fitting strategies} displays the experimental results of the MLP block that utilizes the linear predictor from the DejaVu approach. It is apparent that for models without ReLU activation, such as the LLaMA series, the predictor's recall is marginally lower than that reported in ProSparse (see line \textit{ProSparse} in Table \ref{table: mlp predictor with fitting strategies}), even when a threshold is applied to enhance sparsity. Additionally, the predicted sparsity is also markedly lower than the reference values cited in the literature. The DejaVu's straightforward dual-layer linear predictor struggles to precisely forecast the active neurons in the MLP, and the low levels of predicted sparsity do not translate into a notable increase in computational speed.

\begin{table}[h]
\begin{threeparttable}
    \centering
    \scalebox{0.7}{
    \begin{tabular}{l|c|ccc}
    \toprule
    \textbf{Strategy} & \textbf{Real Sparsity} & \textbf{Activation Recall} & \textbf{Predicted Sparsity} & \textbf{Sparsity Delta} \\
    \midrule
    \textbf{LLaMA-2-7B} & 67.12\% & 86.17 & 29.01\% & 38.11\% \\
    --large batch & 67.12\% & 84.39 & 30.77\% & 36.35\% \; (1.76\%↓) \\
    --focal loss & 67.12\% & 85.90 & 29.40\% & 37.72\% \; (0.39\%↓) \\
    --topk & 65\% & 48.74 & 65.00\% & -  \\  
    \midrule
    \textbf{LLaMA-3-8B} & 45.84\% & 94.19 & 13.08\% & 32.77\% \\
    --topk & 45\% & 67.22 & 44.99\% & -  \\
    \midrule
    \textbf{ProSparse} & 89.32\% & 92.34 & 78.75\% & 10.57\% \\
    \bottomrule
    \end{tabular}
    }
    \begin{tablenotes}\footnotesize
    \item \textbf{Notes}: "TopK" refers to the activation of neurons with the largest $(1-K)$\% of logits.
    \end{tablenotes}
    \caption{Fitting capability of the predictor across different strategies}
    \label{table: mlp predictor with fitting strategies}
\end{threeparttable}
\end{table}

To improve the predictor's fitting ability and mitigate the influence of different experimental configurations, we have conducted a series of additional experiments, the results of which are reported in Table \ref{table: mlp predictor with fitting strategies}. 
These experiments assessed the effects of various enhancements on the model's fitting capacity, such as adjusting the batch size (refer to the line labeled \textit{large batch}) and improving the loss function (refer to the line labeled \textit{focal loss}). The trade-offs remain consistent even after altering the selection strategy. While topK strategies effectively increase the predicted sparsity, they also lead to a substantial decrease in activation recall. The experimental findings reveal a clear trade-off between the predictor's recall and the predicted sparsity, indicating that the double-layer linear structure of the predictor may be nearing its theoretical performance ceiling.

Through our experiment, we find out that the linear predictor encounters difficulties when forecasting the behavior of the vanilla LLaMA model. The experiments detailed in this section confirm a trade-off between the predictor's recall and the level of sparsity it predicts. Experiments reported in Table \ref{table: trade off winno} were designed to examine the effects of solely enhancing the predicted sparsity on the predictor's recall and the overall end-to-end performance. With a defined level of predicted sparsity implemented by the predictor, the average recall outcomes for the first 8 layers and the respective sparsity performance on the lm-eval winogrande task are reported. And the findings indicate that exceeding 30\% sparsity in the vanilla LLaMA-2-7B model significantly degrades its performance, leading to the conlucsion that the linear predictor is not adept at predicting the activation of LLaMA neurons without ReLU activation. 

\begin{table}[h]
\scalebox{0.85}{
\begin{tabular}{l|ccccccc}
\toprule
\textbf{Sparsity} &
  \textbf{Dense} &
  \textbf{5\%} &
  \textbf{6\%} &
  \textbf{10\%} &
  \textbf{15\%} &
  \textbf{30\%} &
  \textbf{50\%} \\
  \midrule
\textbf{Predictor Recall} &
  100\% &
  97.35\% &
  96.02\% &
  94.29\% &
  90.01\% &
  79.93\% &
  68.92\% \\
\textbf{Winogrand(Acc)} &
  74.03 &
  71.03 &
  - &
  68.03 &
  66.93 &
  57.3 &
  - \\
  \bottomrule
\end{tabular}
}
\caption{Trade-off between predictor recall and LLaMA-2-7B performance}
\label{table: trade off winno}
\end{table}

The experiments from this sections demonstrate that the dual-layer linear predictor, as proposed by DejaVu, is not effective for predicting neurons that require dynamic activation in non-ReLU models, such as vanilla LLaMA series. However, the literature mentions that the design of the MLP block predictor should mirror the MLP block's architecture. Consequently, in this section, we adopt the LLaMA MLP structure as the predictor for our experiments. 

\begin{table}[h]
\begin{threeparttable}
    \centering
    \scalebox{0.68}{
    \begin{tabular}{l|c|ccc}
    \toprule
    \textbf{Predictor Structure} & \textbf{Real Sparsity} & \textbf{Predictor Recall} & \textbf{Predicted Sparsity} & \textbf{Sparsity Delta} \\
    \midrule
    \textbf{Linear ReLU} & 67.12\% & 86.17 & 29.01\% & 38.11\% \\
    \midrule
     --LLaMA MLP & 67.12\% & 84.62 & 36.28\% & 30.84\% \; (7.27\%↓) \\
    \multirow{2}*{--LLaMA MLP topK} & 46\% & 68.78 & 46\% & -  \\
    ~ & 50\% & 66.51 & 50\% & - \\
    --w/ LLaMA weight & 50\% & 77.59 & 50\% & - \\
    \midrule
    Literature reference values &65\% & $\sim$80 & $\sim$45\% & $\sim$15\% \\
    \bottomrule
    \end{tabular}
    }
    \begin{tablenotes}\footnotesize
    \item \textbf{Notes}: "TopK" refers to the activation of neurons with the largest $(1-K)$\% of logits.
    \end{tablenotes}
    \caption{Complex predictor's performance on LLaMA-2-7B}
    \label{table: heavy predictor}
\end{threeparttable}
\end{table}

The experiments detailed in Table \ref{table: heavy predictor} indicate that transitioning from a dual-layer linear model to a LLaMA MLP configuration in the predictor's architecture results in an approximate 8\% increase in predicted sparsity (see the line labeled \textit{LLaMA MLP}). Additionally, changing the selection strategy from sigmoid binary classification to a topK approach significantly enhances the predicted sparsity, but this also leads to a noticeable decline in predictor recall (see the lines labeled \textit{LLaMA MLP topK}). This reaffirms the notion that models employing activation functions other than ReLU face significant trade-offs in dynamic activation schemes. Moreover, incorporating the weight of the down projection from the LLaMA MLP block into the predictors allows the model to mirror the theoretical outcomes reported in the literature (see the lines labeled \textit{w/ LLaMA weight}).

\subsubsection{Predictors in Attentions}
Regarding the attention block, since LLaMA-3 utilizes GQA, our experimentation was limited to LLaMA-2. Table \ref{table: att predictor} displays the evaluation results of the attention predictor employing various strategies on the lm-eval schema. "LLaMA-2-7B" denotes the baseline performance of the vanilla LLaMA-2-7B model on lm-eval. "Head mask only" refers to the experimental method outlined in Section \ref{head truncation}, which involves using thresholds determined through CETT search to sparsely and dynamically activate heads without the use of a predictor. The term "sparsity" refers to the manual setting of head sparsity, which dictates the number of active heads per layer.

The experimental results indicate that the sparsity generated by the predictor is below 40\%. The model's performance starts to deteriorate significantly once the sparsity of the attention heads surpasses 5\%, suggesting an exponential decay relationship between head sparsity and model performance, which indicates that implementing a two-layer linear predictor for dynamic sparsity within the attention heads can have a detrimental impact on the model's performance.

\begin{table}[h]
\centering
\scalebox{0.68}{
\begin{tabular}{l|cccccc}
\toprule
  \textbf{Sparsity Strategy} & \textbf{Predicted Sparsity} & \textbf{MMLU} & \textbf{TruthfulQA} & \textbf{Winogrande} & \textbf{GSM8K} & \textbf{Average} \\
  \midrule
  \textbf{LLAMA-2-7B}     & Dense       & 46.87 & 61.24 & 74.03 & 14.48 & 65.18 \\
  \midrule
  --head mask only  & 44\%   & 44.63 & 61.65 & 73.24 & 10.38 & 64.34 \\
  --w/ predictor   & 37.96\% & 27.45 & 60.80 & 61.88 & -     & 51.99 \\
  \midrule
  \multirow{4}*{--topK} & 5\%     & 43.08 & 61.36 & 73.88 & 11.75 & 64.11 \\
  ~  & 30\%    & 24.53 & 60.28 & 61.25 & -     & 53.27 \\
  ~  & 44\%    & 23.67 & 57.60 & 50.91 & -     & 45.75 \\
  ~  & 70\%    & 23.34 & 51.28 & 51.78 & -     & 39.19 \\
  \bottomrule
  \end{tabular}
  }
  \begin{tablenotes}\footnotesize
    \item \textbf{Notes}: "TopK" refers to the activation of neurons with the largest $(1-K)$\% of logits.
    \end{tablenotes}
  \caption{LLaMA-2-7B with attention predictor employing various strategies}
  \label{table: att predictor}
\end{table}

This section elaborates on experiments that investigate dynamic activation through the use of predictors in both MLP and attention blocks. These experiments employ two types of predictors: a dual-layer linear predictor and a predictor with a more intricate structure. The findings reveal that the dual-layer linear predictor has difficulty in learning non-ReLU activation patterns. For both the MLP and attention blocks, a clear trade-off emerges between recall and predicted sparsity. With the dual-layer linear predictor, the model manages to attain only 5\% in predicted sparsity without compromising performance.

Enhancing the complexity of the predictor's structure enables it to attain values that are closer to those reported in the literature in terms of both recall and predicted sparsity. This implies that a sophisticated predictor is more adept at capturing the nuances of dynamic activation, which is vital for preserving model performance while reducing the number of active neurons. Nonetheless, the advantages of a complex predictor are accompanied by drawbacks. The intricate architecture of the predictor imposes a computational load nearly equivalent to that of the vanilla LLaMA, posing challenges for realizing the benefits of dynamic activation.The increased computational burden resulting from this complexity can considerably slow down the model's inference speed and diminish the potential of dynamic sparsity to expedite computations.

In summary, while enhancing the predictor's structure can improve its ability to predict dynamic activation accurately, we must carefully balance the trade-offs between accuracy, computational efficiency, and resource utilization. This balance is essential for the widespread adoption and sustainable use of LLMs like LLaMA-2-7B in various applications.

\section{Conclusion}

In conclusion, our thorough investigation into dynamic activation mechanisms within the LLaMA family of language models has yielded critical insights. While the allure of reduced computational demands and swifter processing through ReLU-based models is undeniable, our empirical research highlights the shortcomings of current dynamic activation techniques. Our extensive testing across a spectrum of sparsity strategies has shown that LLaMA models often fall short of the performance benchmark set by their ReLU-activated counterparts, especially when a high sparsity ratio is required.

The root causes of these performance gaps can be traced to several key issues: the complexity involved in dynamically modulating activation patterns in real-time, the suboptimal sparsity achieved by non-ReLU activation functions, and the loss of crucial information due to KV cache skipping. Our findings not only pinpoint the challenges facing dynamic activation in large-scale LLaMA models but also pave the way for future enhancements. It is our hope that this research will serve as a catalyst for the development of more robust and efficient dynamic activation schemes, ultimately leading to language models that are both computationally efficient and highly effective.

Future advancements in the field of dynamic activation are poised to concentrate on augmenting the sparsity within LLMs, devising algorithms for efficient dynamic and sparse prediction, and establishing a robust training and evaluation framework tailored for sparse predictors. DS-MoE\cite{pan2024dense} presents an innovative training framework designed to tackle the problem of low parameter efficiency commonly found in traditional Mixture-of-Experts (MoE) language models. This framework utilizes dense computation in the training phase and sparse computation in the inference phase, allowing the DS-MoE model to achieve performance on par with dense models while activating only 30\%-40\% of the parameters. Additionally, the paper investigates methods to boost model performance further by incorporating Mutual Information (MI) loss and employing Mixture of Attention (MoA) heads. Specifically, JetMoE\cite{shen2024jetmoe} has adeptly expanded the dynamic activation concept to the attention mechanism, pioneering a sparsely-gated Mixture-of-Experts (SMoE) architecture in attention blocks.

Research in this domain may also involve developing new algorithms that predict which keys and values are most likely to be reused and caching them preferentially. Alternatively, it could explore adaptive caching mechanisms that adjust the stored data based on the current context or task demands. Achieving this level of sophistication in KV cache management would mark a significant milestone in the development of sparsity-aware LLMs, potentially leading to models that are both more efficient and more effective at handling a wide range of language processing tasks.

With an eye towards sparsity enhancement, there is a fertile ground for exploration and creation of novel learning algorithms that can more effectively realize and exploit sparsity. Our future work would encompass the refinement of regularization techniques, or the evolution of predictor structure, and the optimization of schemes designed to evaluation dynamic and sparse predictions, thereby contributing to the overarching quest for computational efficiency and model performance optimization in the field of artificial intelligence.

\bibliographystyle{unsrt}
\bibliography{ref}

\clearpage

\appendix
\section*{Appendix} \label{app:A}
\section{Layer-specific Thresholds under Different CETTs} \label{app: app table}

\begin{table}[h]
\centering
\scalebox{0.6}{
\begin{tabular}{c|c|c|ccccccccccc}
\toprule
\textbf{Model} & \textbf{CETT} & \textbf{Sparsity(\%)} & \textbf{Layer\#1} & \textbf{2} & \textbf{3} & \textbf{4} & \textbf{5} & \textbf{6} & \textbf{7} & \textbf{8} & \textbf{9} & \textbf{10} & \textbf{11} \\
\midrule
\multirow{8}{*}{\textbf{LLaMA-2-7B}} & 0.01 & 14.73 & 0.0002 & 0.0005 & 0.0006 & 0.0009 & 0.0015 & 0.0018 & 0.0022 & 0.0026 & 0.0028 & 0.0031 & 0.0033 \\
 & 0.02 & 21.57 & 0.0004 & 0.0008 & 0.0010 & 0.0016 & 0.0024 & 0.0030 & 0.0037 & 0.0043 & 0.0047 & 0.0051 & 0.0056 \\
 & 0.04 & 31.23 & 0.0007 & 0.0014 & 0.0018 & 0.0028 & 0.0042 & 0.0051 & 0.0062 & 0.0072 & 0.0079 & 0.0085 & 0.0093 \\
 & 0.1 & 49.27 & 0.0016 & 0.0031 & 0.0040 & 0.0059 & 0.0087 & 0.0105 & 0.0128 & 0.0148 & 0.0163 & 0.0175 & 0.0191 \\
 & 0.2 & 67.08 & 0.0032 & 0.0062 & 0.0076 & 0.0110 & 0.0159 & 0.0188 & 0.0228 & 0.0263 & 0.0290 & 0.0311 & 0.0339 \\
 & 0.3 & 78.55 & 0.0053 & 0.0103 & 0.0117 & 0.0168 & 0.0234 & 0.0273 & 0.0330 & 0.0378 & 0.0417 & 0.0449 & 0.0488 \\
 & 0.4 & 86.47 & 0.0085 & 0.0159 & 0.0167 & 0.0235 & 0.0320 & 0.0366 & 0.0439 & 0.0503 & 0.0554 & 0.0594 & 0.0647 \\
 & 0.5 & 92.05 & 0.0143 & 0.0250 & 0.0233 & 0.0321 & 0.0425 & 0.0474 & 0.0565 & 0.0646 & 0.0710 & 0.0762 & 0.0828 \\
\multirow{8}{*}{\textbf{LLaMA-3-8B}} & 0.01 & 14.1 & 0.0002 & 0.0002 & 0.0004 & 0.0005 & 0.0007 & 0.0009 & 0.0010 & 0.0010 & 0.0010 & 0.0010 & 0.0012 \\
\midrule
 & 0.02 & 21.04 & 0.0004 & 0.0004 & 0.0007 & 0.0009 & 0.0012 & 0.0015 & 0.0016 & 0.0017 & 0.0018 & 0.0018 & 0.0020 \\
 & 0.04 & 30.72 & 0.0007 & 0.0008 & 0.0012 & 0.0015 & 0.0021 & 0.0024 & 0.0027 & 0.0029 & 0.0031 & 0.0031 & 0.0033 \\
 & 0.1 & 49.08 & 0.0016 & 0.0018 & 0.0025 & 0.0032 & 0.0043 & 0.0050 & 0.0056 & 0.0059 & 0.0062 & 0.0062 & 0.0067 \\
 & 0.2 & 67.29 & 0.0031 & 0.0033 & 0.0047 & 0.0059 & 0.0076 & 0.0089 & 0.0099 & 0.0104 & 0.0109 & 0.0110 & 0.0118 \\
 & 0.3 & 78.97 & 0.0048 & 0.0050 & 0.0071 & 0.0087 & 0.0110 & 0.0129 & 0.0143 & 0.0149 & 0.0158 & 0.0158 & 0.0171 \\
 & 0.4 & 86.91 & 0.0069 & 0.0070 & 0.0099 & 0.0118 & 0.0148 & 0.0174 & 0.0192 & 0.0199 & 0.0211 & 0.0212 & 0.0228 \\
 & 0.5 & 92.47 & 0.0099 & 0.0095 & 0.0135 & 0.0157 & 0.0192 & 0.0228 & 0.0250 & 0.0258 & 0.0276 & 0.0276 & 0.0295 \\
 \toprule
\textbf{Model} & \textbf{CETT} & \textbf{Sparsity(\%)} & \textbf{Layer\#12} & \textbf{13} & \textbf{14} & \textbf{15} & \textbf{16} & \textbf{17} & \textbf{18} & \textbf{19} & \textbf{20} & \textbf{21} & \textbf{22} \\
\midrule
\multirow{8}{*}{\textbf{LLaMA-2-7B}} & 0.01 & 14.73 & 0.0035 & 0.0037 & 0.0042 & 0.0044 & 0.0049 & 0.0059 & 0.0059 & 0.0063 & 0.0068 & 0.0071 & 0.0070 \\
 & 0.02 & 21.57 & 0.0059 & 0.0062 & 0.0069 & 0.0073 & 0.0083 & 0.0098 & 0.0099 & 0.0106 & 0.0114 & 0.0120 & 0.0119 \\
 & 0.04 & 31.23 & 0.0099 & 0.0104 & 0.0117 & 0.0125 & 0.0140 & 0.0164 & 0.0168 & 0.0181 & 0.0193 & 0.0204 & 0.0203 \\
 & 0.1 & 49.27 & 0.0203 & 0.0215 & 0.0240 & 0.0256 & 0.0288 & 0.0336 & 0.0347 & 0.0375 & 0.0397 & 0.0427 & 0.0425 \\
 & 0.2 & 67.08 & 0.0359 & 0.0383 & 0.0427 & 0.0458 & 0.0516 & 0.0599 & 0.0622 & 0.0676 & 0.0717 & 0.0776 & 0.0775 \\
 & 0.3 & 78.55 & 0.0516 & 0.0552 & 0.0615 & 0.0663 & 0.0748 & 0.0869 & 0.0906 & 0.0989 & 0.1054 & 0.1146 & 0.1146 \\
 & 0.4 & 86.47 & 0.0684 & 0.0732 & 0.0817 & 0.0886 & 0.1004 & 0.1171 & 0.1223 & 0.1343 & 0.1442 & 0.1576 & 0.1578 \\
 & 0.5 & 92.05 & 0.0875 & 0.0937 & 0.1051 & 0.1149 & 0.1314 & 0.1542 & 0.1613 & 0.1791 & 0.1946 & 0.2147 & 0.2151 \\
 \midrule
\multirow{8}{*}{\textbf{LLaMA-3-8B}} & 0.01 & 14.1 & 0.0012 & 0.0013 & 0.0013 & 0.0014 & 0.0015 & 0.0015 & 0.0015 & 0.0015 & 0.0015 & 0.0015 & 0.0015 \\
 & 0.02 & 21.04 & 0.0020 & 0.0023 & 0.0023 & 0.0024 & 0.0024 & 0.0025 & 0.0026 & 0.0024 & 0.0024 & 0.0025 & 0.0026 \\
 & 0.04 & 30.72 & 0.0034 & 0.0038 & 0.0038 & 0.0040 & 0.0042 & 0.0042 & 0.0043 & 0.0041 & 0.0040 & 0.0042 & 0.0043 \\
 & 0.1 & 49.08 & 0.0070 & 0.0079 & 0.0076 & 0.0082 & 0.0084 & 0.0087 & 0.0089 & 0.0083 & 0.0082 & 0.0085 & 0.0088 \\
 & 0.2 & 67.29 & 0.0124 & 0.0139 & 0.0134 & 0.0146 & 0.0149 & 0.0156 & 0.0157 & 0.0148 & 0.0146 & 0.0151 & 0.0156 \\
 & 0.3 & 78.97 & 0.0179 & 0.0200 & 0.0192 & 0.0212 & 0.0215 & 0.0229 & 0.0228 & 0.0215 & 0.0214 & 0.0221 & 0.0227 \\
 & 0.4 & 86.91 & 0.0239 & 0.0267 & 0.0255 & 0.0287 & 0.0288 & 0.0313 & 0.0309 & 0.0292 & 0.0291 & 0.0300 & 0.0308 \\
 & 0.5 & 92.47 & 0.0311 & 0.0349 & 0.0329 & 0.0380 & 0.0377 & 0.0423 & 0.0411 & 0.0390 & 0.0388 & 0.0400 & 0.0410 \\
 \toprule
\textbf{Model} & \textbf{CETT} & \textbf{Sparsity(\%)} & \textbf{Layer\#23} & \textbf{24} & \textbf{25} & \textbf{26} & \textbf{27} & \textbf{28} & \textbf{29} & \textbf{30} & \textbf{31} & \textbf{32} & \textbf{} \\
\midrule
\multirow{8}{*}{\textbf{LLaMA-2-7B}} & 0.01 & 14.73 & 0.0072 & 0.0076 & 0.0074 & 0.0082 & 0.0085 & 0.0096 & 0.0107 & 0.0125 & 0.0176 & 0.0410 &  \\
 & 0.02 & 21.57 & 0.0122 & 0.0128 & 0.0127 & 0.0139 & 0.0143 & 0.0161 & 0.0181 & 0.0211 & 0.0297 & 0.0704 &  \\
 & 0.04 & 31.23 & 0.0210 & 0.0219 & 0.0218 & 0.0237 & 0.0246 & 0.0274 & 0.0310 & 0.0358 & 0.0508 & 0.1219 &  \\
 & 0.1 & 49.27 & 0.0441 & 0.0457 & 0.0460 & 0.0497 & 0.0516 & 0.0570 & 0.0648 & 0.0743 & 0.1074 & 0.2542 &  \\
 & 0.2 & 67.08 & 0.0808 & 0.0832 & 0.0842 & 0.0905 & 0.0940 & 0.1035 & 0.1183 & 0.1361 & 0.2013 & 0.4710 &  \\
 & 0.3 & 78.55 & 0.1200 & 0.1232 & 0.1249 & 0.1339 & 0.1388 & 0.1530 & 0.1761 & 0.2035 & 0.3065 & 0.7272 &  \\
 & 0.4 & 86.47 & 0.1659 & 0.1700 & 0.1724 & 0.1847 & 0.1906 & 0.2106 & 0.2452 & 0.2848 & 0.4303 & 1.2104 &  \\
 & 0.5 & 92.05 & 0.2274 & 0.2325 & 0.2352 & 0.2521 & 0.2574 & 0.2859 & 0.3387 & 0.3922 & 0.5866 & 2.1083 &  \\
 \midrule
\multirow{8}{*}{\textbf{LLaMA-3-8B}} & 0.01 & 14.1 & 0.0014 & 0.0013 & 0.0014 & 0.0018 & 0.0021 & 0.0029 & 0.0032 & 0.0043 & 0.0068 & 0.0189 \\
 & 0.02 & 21.04 & 0.0023 & 0.0022 & 0.0024 & 0.0031 & 0.0035 & 0.0048 & 0.0052 & 0.0073 & 0.0114 & 0.0331 \\
 & 0.04 & 30.72 & 0.0039 & 0.0037 & 0.0040 & 0.0052 & 0.0059 & 0.0079 & 0.0089 & 0.0123 & 0.0194 & 0.0609 \\
 & 0.1 & 49.08 & 0.0079 & 0.0074 & 0.0081 & 0.0106 & 0.0120 & 0.0163 & 0.0179 & 0.0258 & 0.0411 & 0.1417  \\
 & 0.2 & 67.29 & 0.0140 & 0.0131 & 0.0143 & 0.0190 & 0.0214 & 0.0299 & 0.0325 & 0.0491 & 0.0783 & 0.3026  \\
 & 0.3 & 78.97 & 0.0203 & 0.0188 & 0.0206 & 0.0283 & 0.0316 & 0.0460 & 0.0486 & 0.0786 & 0.1163 & 0.6145  \\
 & 0.4 & 86.91 & 0.0273 & 0.0251 & 0.0276 & 0.0397 & 0.0439 & 0.0686 & 0.0685 & 0.1161 & 0.1534 & 0.9915  \\
 & 0.5 & 92.47 & 0.0359 & 0.0328 & 0.0362 & 0.0568 & 0.0613 & 0.1089 & 0.0942 & 0.1705 & 0.2068 & 1.4966  \\
 \bottomrule
\end{tabular}
}
\caption{Layer-wise threshold for LLaMA models}
\label{table: cett layer wise}
\end{table}

\end{document}